\documentclass[sigconf]{acmart}

\AtBeginDocument{%
  }

\copyrightyear{2025}
\acmYear{2025}
\setcopyright{acmlicensed}
\acmConference[ICMR '25]{Proceedings of the 2025
International Conference on Multimedia Retrieval}{June 30-July 3,
2025}{Chicago, IL, USA}
\acmBooktitle{Proceedings of the 2025 International Conference on Multimedia
Retrieval (ICMR '25), June 30-July 3, 2025, Chicago, IL, USA}
\acmDOI{10.1145/3731715.3733348}
\acmISBN{979-8-4007-1877-9/2025/06}

\usepackage{multirow}
\usepackage{marvosym}
\usepackage{pifont}
\usepackage{utfsym}
\usepackage{dsfont}
\usepackage{balance}
\usepackage{amssymb}
\begin{document}

\title{Generative Emotion Cause Explanation in Multimodal Conversations}


\author{Lin Wang}
\email{q3222345200@gmail.com}
\affiliation{%
  \institution{Northeastern University}
  \city{Shenyang}
  \state{Liaoning}
  \country{China}
}

\author{Xiaocui Yang}
\email{yangxiaocui@cse.neu.edu.cn}
\affiliation{%
  \institution{Northeastern University}
  \city{Shenyang}
  \state{Liaoning}
  \country{China}
}

\author{Shi Feng}
\email{fengshi@cse.neu.edu.cn}
\affiliation{%
  \institution{Northeastern University}
  \city{Shenyang}
  \state{Liaoning}
  \country{China}
}
\authornote{Corresponding author}

\author{Daling Wang}
\email{wangdaling@cse.neu.edu.cn}
\affiliation{%
\institution{Northeastern University}
  \city{Shenyang}
  \state{Liaoning}
  \country{China}
}

\author{Yifei Zhang}
\email{zhangyifei@cse.neu.edu.cn}
\affiliation{%
\institution{Northeastern University}
  \city{Shenyang}
  \state{Liaoning}
  \country{China}
}

\author{Zhitao Zhang}
\email{windowschuang@163.com}
\affiliation{
\institution{Shenyang Women's and Children's Hospital}
    \city{Shenyang}
    \state{Liaoning}
    \country{China}
}

\begin{abstract}
Multimodal conversation, a crucial form of human communication, carries rich emotional content, making the exploration of the causes of emotions within it a research endeavor of significant importance. However, existing research on the causes of emotions typically employs an utterance selection method within a single textual modality to locate causal utterances. This approach remains limited to coarse-grained assessments, lacks nuanced explanations of emotional causation, and demonstrates inadequate capability in identifying multimodal emotional triggers. Therefore, we introduce a task—\textbf{Multimodal Emotion Cause Explanation in Conversation (MECEC)}. This task aims to generate a summary based on the multimodal context of conversations, clearly and intuitively describing the reasons that trigger a given emotion. To adapt to this task, we develop a new dataset (ECEM) based on the MELD dataset. ECEM combines video clips with detailed explanations of character emotions, helping to explore the causal factors behind emotional expression in multimodal conversations. A novel approach, FAME-Net, is further proposed, that harnesses the power of Large Language Models (LLMs) to analyze visual data and accurately interpret the emotions conveyed through facial expressions in videos. By exploiting the contagion effect of facial emotions, FAME-Net effectively captures the emotional causes of individuals engaged in conversations. Our experimental results on the newly constructed dataset show that FAME-Net outperforms several excellent baselines. Code and dataset are available at \url{https://github.com/3222345200/FAME-Net}.
\end{abstract}



\maketitle

\section{Introduction}

Multimodal conversation plays a significant role in our daily life, influencing our thoughts and behaviors. Research \cite{woo2020convergence, agnew2014interpersonal, costa2013emotional} demonstrates that exploring the causes of emotions holds great importance, as a deeper understanding of human emotions and their triggers promotes mental health, optimizes interpersonal relationships, and fosters the development of more empathetic and humanized artificial intelligence systems \cite{fei2024empathyear, gao2021improving, li2021towards}. In recent years, researchers have increasingly focused on how to explore the underlying causes of emotions. Consequently, a new task called Emotion Cause Extraction (ECE) \cite{xia2019rthn, zheng2022ueca} is proposed, along with the introduction of a benchmark dataset named RECCON \cite{poria2021recognizing}. The goal of the ECE task is to identify the specific utterances in a conversation that causally trigger the emotion expressed in the target utterance.

\begin{figure}[t]
  \includegraphics[width=\columnwidth]{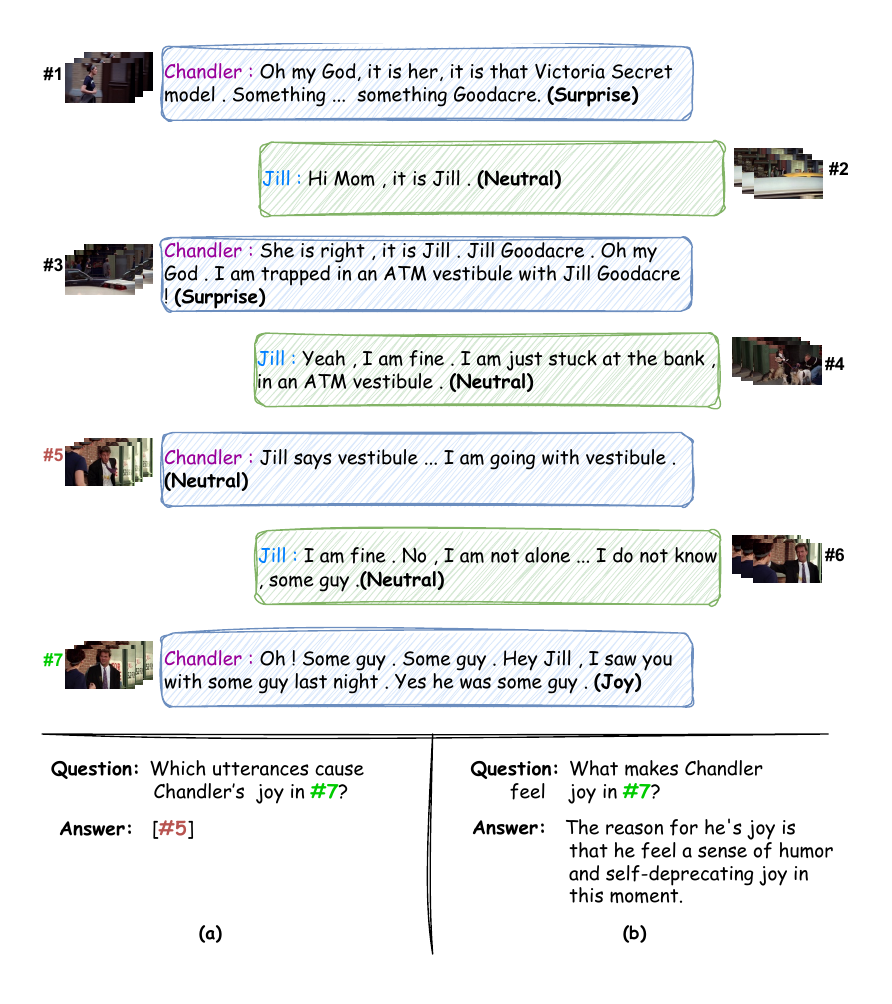}
  \caption{Comparison between selective and generative: \textbf{(a)} Selective task
 requires identifying causal statements (\#7) in the target utterance (\#7) that trigger the speaker's emotions. \textbf{(b)} Generative tasks require extracting the causes for the emotions contained in the target utterances from the conversation history.}
  \label{fig:fig6}
\end{figure}

\citeauthor{wang2022multimodal} \cite{wang2022multimodal} further propose a new task, Multimodal Emotion-Cause Pair Extraction in Conversations (MECPE), which expands the research scope to include multimodal content. However, the identification of emotional causes in this task still relies on selective cause extraction rather than cause generation. Selective cause extraction can only perform a simple selection of emotional causes at a coarse-grained level and fails to capture fine-grained emotional causes that require deeper contextual reasoning. For those emotional utterances, if the causal relationships behind them are primarily conveyed through visual or auditory information, determining the specific causes that trigger emotions becomes particularly challenging. Therefore, generating detailed explanations for emotional causes can provide a more comprehensive and nuanced account of the underlying reasons behind emotions. The comparison between selective and generative approaches is illustrated in Figure~\ref{fig:fig6}.
\citet{wang2024observe} introduce the Multimodal Emotion Cause Generation in Conversations (MECGC) task and the ECGF dataset to address limitations in selective emotion cause discovery. However, with an average explanation length of only 8.64 words, ECGF struggles to capture the complexity of emotion causes, potentially missing crucial contextual details and subtle emotional triggers.

To enable more comprehensive emotion cause analysis, we propose the \textbf{Multimodal Emotion Cause Explanation in Conversation (MECEC)} task. MECEC aims to generate detailed explanations for emotional expressions by leveraging both textual and visual information from conversations. We introduce the \textbf{Emotion Cause Explanation (ECEM)} dataset, built upon MELD \cite{poria2018meld}, comprising 7,273 video clips with rich multimodal annotations. ECEM features an average explanation length of 14.4 words, allowing for more nuanced descriptions of emotional causes.
Our rigorous annotation process involves five graduate students specializing in sentiment analysis. Two annotators independently label each instance, and their annotations are evaluated using BERTScore \cite{zhang2019bertscore}. When the BERTScore exceeds 0.75, the remaining three students vote to determine the final annotation. Otherwise, annotators discuss and revise until a consensus is reached. This protocol, enhanced by automatic evaluation metrics, ensures higher annotation reliability compared to ECGF's approach, which lacks systematic consistency checks between annotators.

To address the MECEC task, we propose FAME-Net (Facial-Aware Multimodal Emotion Explanation Network), a novel architecture built upon LLaVA \cite{NEURIPS2023_6dcf277e}. Drawing from research showing that facial expressions provide crucial cues for emotion cause identification \cite{hatfield1993emotional, ekman1971constants}, FAME-Net incorporates both visual and facial emotion recognition to capture subtle emotional dynamics often missed by text-only analysis.
FAME-Net features a specialized two-stage facial emotion recognition pipeline, first detecting and tracking faces within video scenes, then extracting facial emotion features through multi-scale feature networks and global depth convolution techniques. This comprehensive multimodal approach enables FAME-Net to identify both explicit and implicit emotional triggers, achieving strong performance across standard text generation metrics.

Our main contributions are as follows:
\begin{itemize}
    \item We propose a new Emotion Cause Explanation dataset  (ECEM) based on MELD. This dataset provides detailed natural language explanations for the emotion causes of the target utterances and is specifically designed for the training and evaluation of MECEC tasks.
    \item We propose FAME-Net, integrating visual modality and facial emotion recognition for comprehensive emotional cause analysis. Through a two-stage pipeline, it accurately extracts emotional features, improving multi-modal emotion cause explanation and generation performance.
    \item The experimental results on the ECGF and ECEM datasets demonstrate that our proposed model, FAME-Net, surpasses existing methods across the majority of evaluation metrics.
\end{itemize}

\section{Related Works}

\subsection{Causal Emotion Entailment}

Over the past decade, emotion cause analysis has gained attention for enhancing emotional intelligence, focusing on two key tasks: Emotion Cause Extraction (ECE) and Emotion Cause Pair Extraction (ECPE). Initially proposed by Lee et al. \cite{lee2010text}, ECE identifies text segments explaining emotion causes. Gui et al. \cite{gui2016emotion} later refined ECE at the clause level using a Chinese dataset. Poria et al. \cite{poria2021recognizing} advanced this with dual-stream attention (TSAM) and commonsense integration. While existing tasks often use multiple-choice formats, generative causal reasoning remains underexplored. Xia and Ding \cite{xia2019emotion} proposed ECPE to jointly extract emotions and causes, broadening applications. Wang et al. \cite{wang2022multimodal} introduced Multimodal ECPE (MECPE) by incorporating text, audio, and video. Recently, Wang et al. \cite{wang2024observe} proposed MECGC for fine-grained causal explanations and MECEC, a generative task ensuring conversational continuity in multimodal dialogues. These advancements highlight the field’s evolving potential.

\begin{figure*}[t]

  \includegraphics[width=\linewidth]{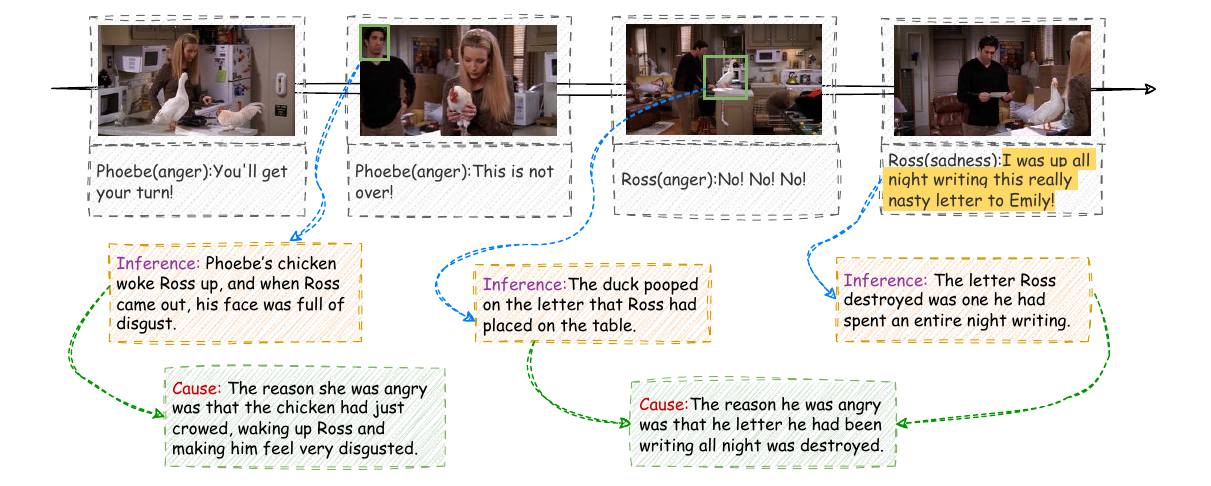}
  \caption {An example of the annotated conversation in our ECEM dataset. 
  Inference represents the reasoning conducted during the annotation process, while cause, represents the emotional causes finally annotated in the dataset.}
\label{fig:fig1}
\end{figure*}
\subsection{Large Language Models}

The advent of Large Language Models (LLMs) has greatly impacted Natural Language Processing (NLP) and multimodal learning, revolutionizing artificial intelligence. LLMs like OpenAI's GPT \cite{brown2020language} excel at a variety of language tasks, including translation, summarization, question answering, and content generation. These models are trained on vast data, allowing them to perform with impressive accuracy and engage in complex conversations and reasoning, previously thought to be human-exclusive. This shift has transformed NLP applications, from research to commercial uses such as chatbots and personalized recommendations.
Simultaneously, models like CLIP \cite{radford2021learning} and BLIP \cite{li2022blip} have advanced the integration of visual and textual understanding. CLIP can process both images and text, enabling tasks like image search and caption generation. BLIP further refines this, advancing visual-textual applications in areas like visual storytelling. Multimodal learning now includes dynamic media like video and audio. Models such as LLaVA \cite{liu2024visual} and Video-ChatGPT \cite{maaz2023video} combine video and audio analysis with AI systems, enhancing interactions and improving applications like video summarization and real-time communication. This shift reflects the growing need for more sophisticated AI that mimics human communication across multiple senses. FAME-Net can take into account the emotional factors of the individuals in the video while processing visual information.

\section{Dataset Construction}

\begin{table*}
  \centering
    \caption{\label{tab:tab1}
    Comparison of Cause Detection Datasets, including traditional Emotion Cause Analysis (ECA), Emotion Cause Detection (ECD), Sarcasm Detection (SD), and Humor Detection (HD), etc.
    The units of the instances include sentence
$(s)$, document $(d)$, post $(p)$, utterance $(u)$ and video $(v)$. T, A, V stands for text, audio and video. }
  \begin{tabular}{cccccccc}
\toprule[1pt]
\multirow{2}{*}{Dataset} & \multicolumn{3}{c}{Modality} & \multirow{2}{*}{Task} & \multirow{2}{*}{Cause type} & \multirow{2}{*}{Scene} & \multirow{2}{*}{Ins} \\ \cline{2-4}
& T    & A    & V   &&&&\\ \midrule[1pt]
ECE Corpus \cite{gui2018event}               & {\color{green}\usym{2713}}     &{\color{red}\usym{2717}}        & {\color{red}\usym{2717}}  & ECE                         & Extractive                  & News                   & 2,105 $d$              \\

NTCIR-13-ECA \cite{gao2017overview}             & {\color{green}\usym{2713}}       & {\color{red}\usym{2717}}        & {\color{red}\usym{2717}}   & ECA                          & Extractive                  & Fiction                & 2,403 $d$              \\

Weibo-Emotion \cite{cheng2017emotion}            & {\color{green}\usym{2713}}       & {\color{red}\usym{2717}}        & {\color{red}\usym{2717}}  & ECD                           & Extractive                  & Blog                   & 7,000 $p$              \\

REMAN \cite{kim2018feels}                    & {\color{green}\usym{2713}}       & {\color{red}\usym{2717}}        & {\color{red}\usym{2717}}    & ECD                         & Extractive                  & Fiction                & 1,720 $d$              \\

GoodNewsEveryone \cite{bostan2019goodnewseveryone}         & {\color{green}\usym{2713}}       & {\color{red}\usym{2717}}        & {\color{red}\usym{2717}}    & ECD                         & Extractive                  & News                   & 5,000 $s$              \\ \hline

MHD \cite{patro2021multimodal} & {\color{green}\usym{2713}}       & {\color{red}\usym{2717}}        & {\color{green}\usym{2713}} & HD  & Extractive & Sitcom & 13,633 $u$ \\

UR-FUNNY \cite{hasan2019ur}  & {\color{green}\usym{2713}}       & {\color{green}\usym{2713}}        & {\color{green}\usym{2713}} &   HD  & Extractive & Speech & 16,514 $u$ \\

MUStARD \cite{castro2019towards}  & {\color{green}\usym{2713}}       & {\color{green}\usym{2713}}        & {\color{green}\usym{2713}} &   SD & Extractive & Sitcom & 6,365 $v$ \\

WITS \cite{kumar2022did} & {\color{green}\usym{2713}}       &  {\color{green}\usym{2713}}        & {\color{green}\usym{2713}} & SD & Generative & Sitcom & 2,240 $u$ \\

SMILE \cite{hyun2023smile}  & {\color{green}\usym{2713}}       & {\color{green}\usym{2713}}        & {\color{green}\usym{2713}} &   HD  & Generative & Sitcom\&Talks & 887 $v$ \\

ExFunTube \cite{ko2023can} & {\color{green}\usym{2713}}       & {\color{green}\usym{2713}}        & {\color{green}\usym{2713}} &   HD  & Generative & Youtube & 10,136 $v$ \\
\hline
RECCON-IE \cite{poria2021recognizing}                & {\color{green}\usym{2713}}       & {\color{red}\usym{2717}}        & {\color{red}\usym{2717}}   & ECE                          & Extractive                  & Conv                   & 665 $u$                \\

RECCON-DD \cite{poria2021recognizing}                & {\color{green}\usym{2713}}       & {\color{red}\usym{2717}}        & {\color{red}\usym{2717}}    & ECE                         & Extractive                  & Conv                   & 11,104 $u$             \\

ConvECPE \cite{li2022ecpec} & {\color{green}\usym{2713}}       & {\color{red}\usym{2717}}        & {\color{red}\usym{2717}} & ECPE & Extractive & Conv & 7,433 $u$ \\

ECF \cite{wang2022multimodal}                      & {\color{green}\usym{2713}}       & {\color{green}\usym{2713}}        & {\color{green}\usym{2713}}    & MECPE                          & Extractive                  & Conv                   & 13,619 $u$             \\

ECGF \cite{wang2024observe}                      & {\color{green}\usym{2713}}       & {\color{green}\usym{2713}}        & {\color{green}\usym{2713}}    & MECGC                          & Generative                  & Conv                   & 13,619 $u$             \\

\textbf{ECEM (ours)  }             & {\color{green}\usym{2713}}       & {\color{green}\usym{2713}}        & {\color{green}\usym{2713}}     & MECEC                         & Generative                  & Conv                   & 7,273 $u$               \\ \bottomrule[1pt]
  \end{tabular}
\end{table*}

\subsection{Annotation Program}
We first provide a detailed explanation of the dataset annotation method. To ensure consistency in style and accuracy of the annotated data, we refer to the research findings \cite{lin2014microsoft, fang2015captions}. Specifically, we invite five graduate students specializing in sentiment analysis to participate in annotation work. We first train them on annotation methods, ensuring they correctly understand the task definition and proficiently master the use of annotation tools, thereby guaranteeing the quality of the dataset we construct. Then, two students independently annotated the entire dataset. We calculate the BERTScore \cite{zhang2019bertscore} of their annotations. For cases with scores exceeding 0.75, the remaining three graduate students independently evaluate the annotations, with the final result determined by majority vote. When scores are below 0.75, the annotators discuss discrepancies to reach consensus. This ensures the rationality and consistency of data annotation. During the data annotation process, we observe issues such as missing textual data relative to the video content and incorrect use of punctuation. Therefore, in our experiments, we used a combination of WhisperX \cite{bain2023whisperx} and WhisperAT \cite{gong2023whisper} to transcribe audio information from videos to correct the original data and improve data quality.

\paragraph{Data Pre-Processing} We preprocess the MELD dataset by merging preceding video clips into a continuous sequence to accurately capture emotional causes of each utterance without interference from future content. This method is beneficial for elucidating the emotional causes associated with the final discourse in the video segments. For example, in a dialogue consisting of two utterances, \mbox{\(U_0\)} and \mbox{\(U_1\)}, the concatenated results would be \mbox{\((U_0)\)} and \mbox{\((U_0, U_1)\)}. During the concatenation process, overlapping segments between the original video clips resulted in redundant content. Therefore, we design an algorithm to detect and remove duplicate frames to enhance the quality of our dataset.

\paragraph{Annotation Guidelines} We focus more on different emotions; therefore, this paper does not involve the neutral emotions. Before the formal annotation, we randomly assign 100 videos to annotators for training, and formal annotation only begins after the training results receive unanimous approval from evaluators. For each non-neutral utterance, annotators are required to find corresponding evidence from the three modalities of the dialogue and briefly describe their reasoning in one to three sentences. We specifically request that annotators pay close attention to the video and audio modalities to ensure the accuracy and completeness of the reasons provided. Annotated examples are shown in the Figure~\ref{fig:fig1}.

\subsection{Data Statistics and Analysis}

\begin{figure}[t]
  \includegraphics[width=\columnwidth]{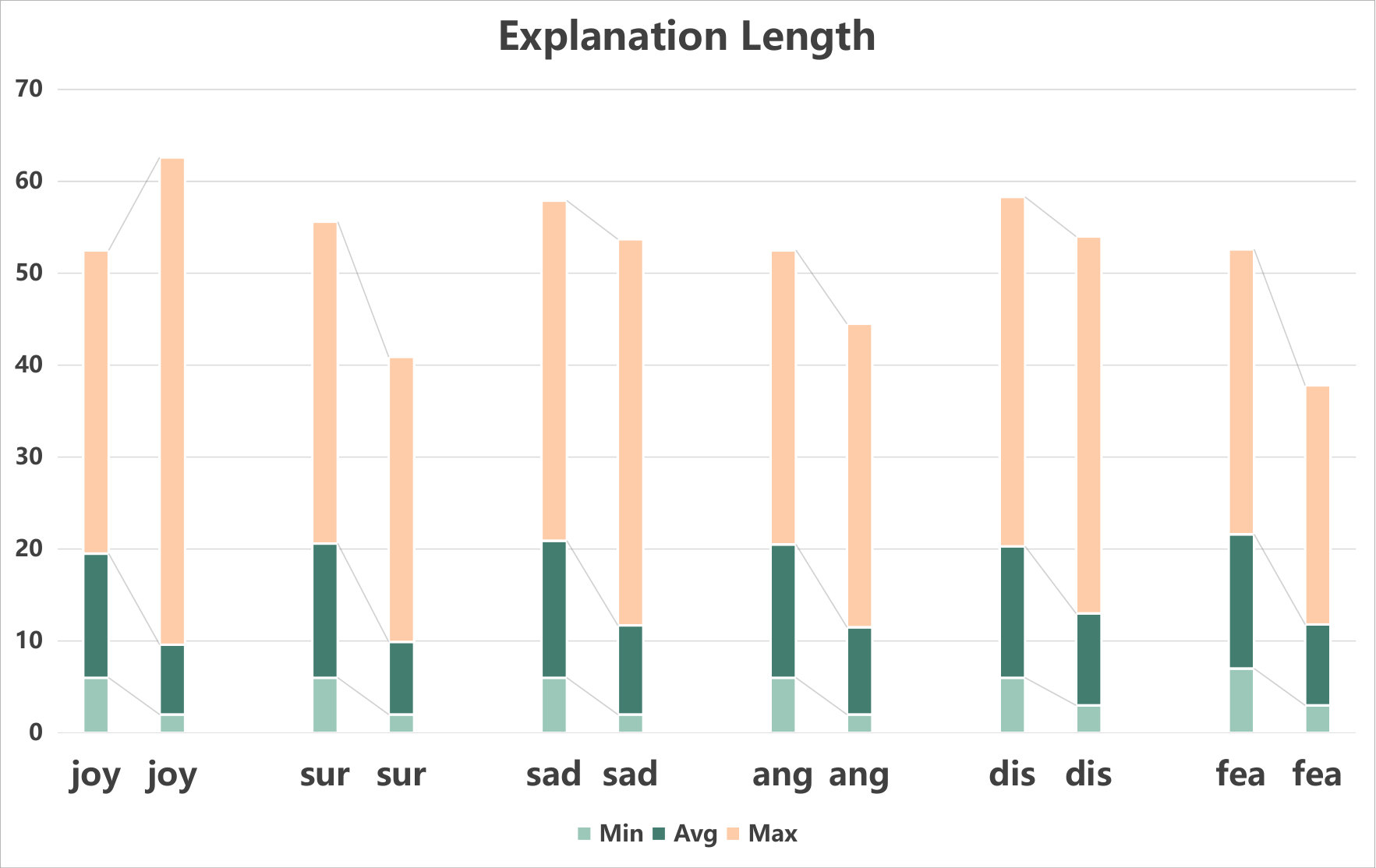}
  \caption{Statistical analysis of the reason length for emotion causes is conducted on the datasets ECEM and ECGF. The left side represents ECEM, while the right side represents ECGF.}
  \label{fig:fig4}
\end{figure}

\begin{figure}[h]
  \includegraphics[width=0.88\columnwidth]{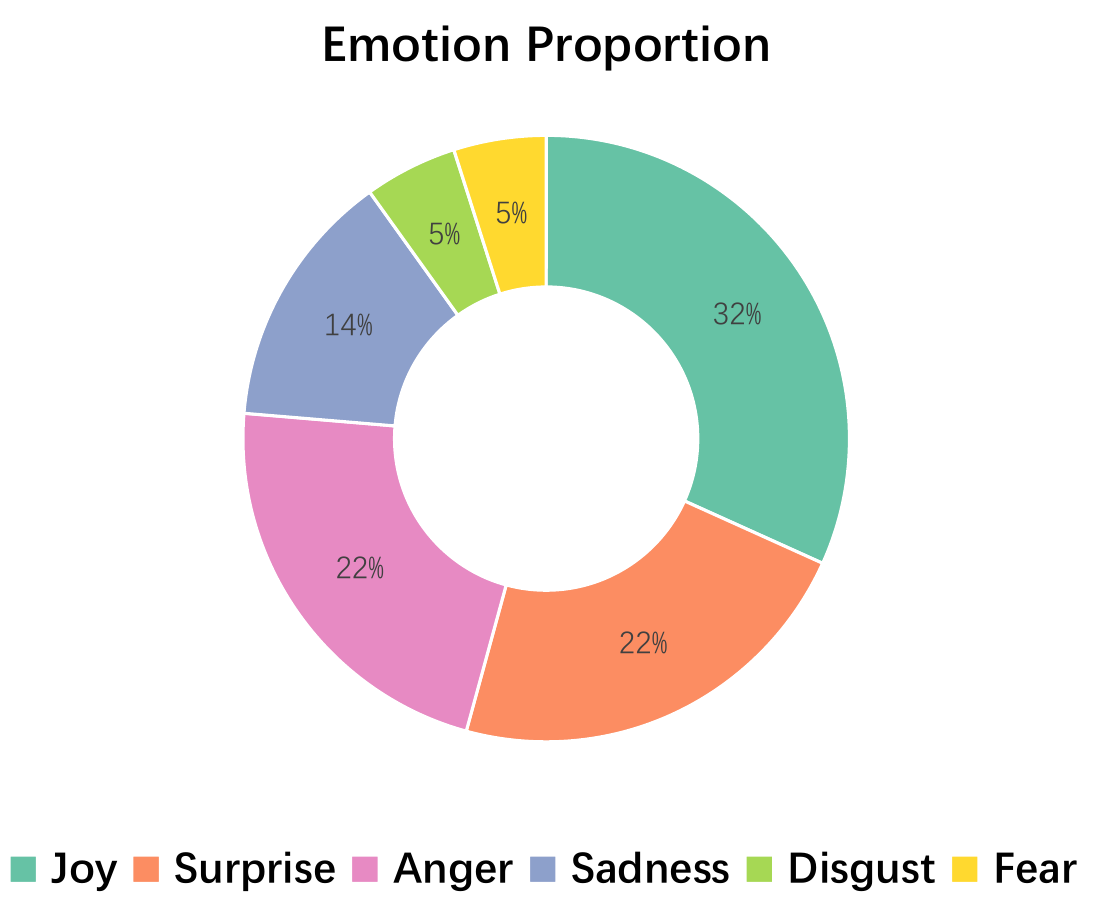}
  \caption{Proportion of six classified emotions.}
  \label{fig:fig3}
\end{figure}
\begin{figure*}[t]
  \includegraphics[width=0.85\linewidth]{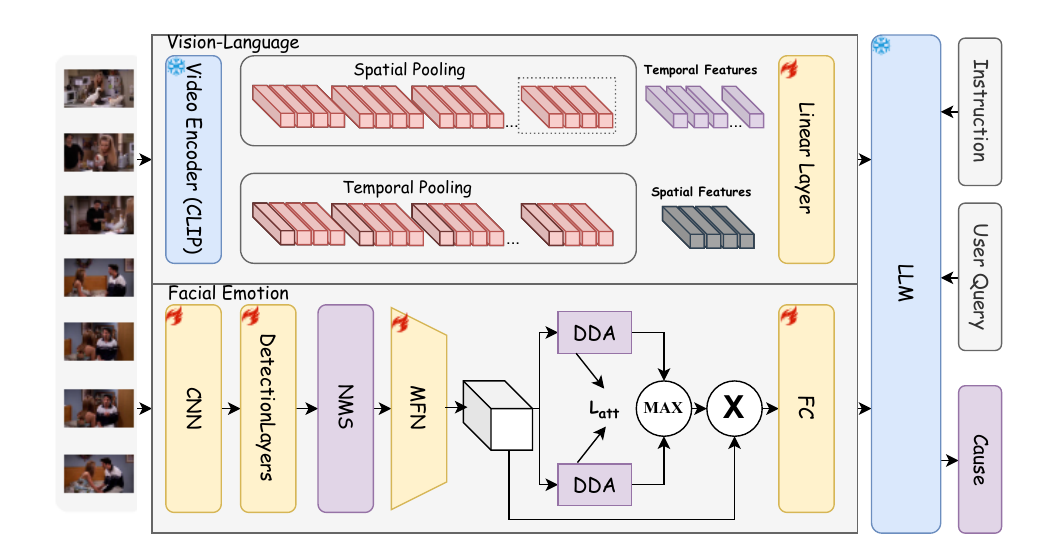}
  \caption {Overview of our proposed FAME-Net. The MFN network includes multiple convolutional layers, depthwise convolutional layers, residual connections, and a final linear layer. DDA includes convolution, attention mechanisms, and linear transformations. X represents multiplication.}
\label{fig:fig5}
\end{figure*}
\paragraph{Comparison with Existing Datasets and Data Analysis}
As shown in Table~\ref{tab:tab1}, by comparing the ECEM dataset with mainstream datasets in the field of emotion cause detection, we find that existing datasets (such as those for humor and sarcasm detection) are often limited to single emotions or fragmented explanations. Although the ECGF \cite{wang2024observe} dataset supports multi-emotion explanations, its annotation method based on video segmentation results in excessively short explanations—averaging 8.64 words. In contrast, the ECEM dataset not only covers six emotions but also achieves more comprehensive emotion attribution through coherent explanations averaging 14.4 words. The comparison of explanation lengths between ECGF and ECEM is illustrated in Figure~\ref{fig:fig4}. To ensure data quality, we remove neutral emotion samples during the construction process and conduct statistical analysis on the proportions of the six emotions, with the results shown in Figure~\ref{fig:fig3}. Additionally, we perform a statistical analysis of the annotation text lengths, revealing that explanations for different emotions exhibit high consistency in terms of maximum, minimum, and average values. This standardized pattern not only validates the rigor of the annotation process but also confirms the dataset's reliability in cross-emotion explanation capabilities. This dual advantage—surpassing the structural limitations of existing datasets while incorporating a rigorous data validation mechanism—makes ECEM the most interpretable benchmark for emotion cause analysis to date.

\section{Method}

\subsection{Overview}

We present the Facial-Aware Multimodal Emotion Explanation Network (FAME-Net), a multimodal large language model (LLM) built on the LLaVA framework \cite{liu2024visual}. LLaVA is selected for its optimal balance between computational efficiency and model performance, as well as its adaptable interfaces for video-related tasks, which allow researchers to prioritize refining temporal modeling and advancing complex reasoning mechanisms. As illustrated in Figure~\ref{fig:fig5}, FAME-Net integrates visual, textual, and facial emotion modalities. Within its visual-language module, spatiotemporal feature fusion is employed to concurrently capture temporal dynamics and spatial structural patterns in video data, enhancing fine-grained content understanding. Facial emotion recognition complements text-based analysis limitations \cite{ekman1971constants}, quantifies emotional contagion to uncover latent affective drivers \cite{hatfield1993emotional}, and detects discrepancies between facial expressions and verbal cues to identify concealed emotional states \cite{chartrand1999chameleon}. Our work focuses on modeling the temporal evolution and contextual dependencies of facial expressions in video streams, proposing a novel architecture tailored for dynamic emotion analysis.

During its development, FAME-Net integrates advanced mechanisms to align visual signals with language processing, enhancing multimodal analysis. The model features a CLIP-based video encoder \cite{radford2021learning} tailored to capture the information from spatial and temporal dimensions of video data, alongside a facial emotion recognition module that accurately identifies and interprets subtle emotional nuances. This combination enables FAME-Net to reveal deeper emotional insights beyond conventional text analysis.

\subsection{Vision-Language Module}

Our architecture, as shown in the "Vision-Language" section of Figure~\ref{fig:fig5}, employs the CLIP ViT-L/14@336 as the visual encoder, adapting it from image processing to video processing to meet the requirements of FAME-Net. This adaptation is crucial for the model to effectively capture spatio-temporal representations in videos. For a video, $V=\{I_1, ..., I_k, ..., I_T\}$, where $I_k \in\mathbb{R}^{ H\times W\times C}$ and  $T$ denotes the number of frames. The encoder independently processes each frame, generating the frame-level embedding $x_k = CLIP(I_k), x_k\in\mathbb{R}^{ h\times w\times D}$, where $h=H / p,\ w=W / p$ and $p$ represents the patch size, and $h \times w = N$. The video embedding can be obtained, $X = [x_1, .., x_k, ..., x_T]$.

To construct a comprehensive video-level representation, we apply average pooling along the temporal dimension of frame-level embeddings, resulting in a temporal representation $R_t\in\mathbb{R}^{T\times D}$:
\begin{equation}
R_t=\frac{1}{N}\sum_{i=1}^{N}X
\end{equation}
This temporal pooling technique effectively integrates information across frames. Similarly, we apply average pooling along the spatial dimension to obtain the spatial representation $R_s\in\mathbb{R}^{N\times D}$:

\begin{equation}
R_s=\frac{1}{T}\sum_{i=1}^{T}X
\end{equation}
The final video-level features $v_j$ are a concatenation of these temporal and spatial features.
\begin{equation}
R=[R_t \oplus  R_s]\in\mathbb{R}^{(T+N)\times D}
\label{eq:2}
\end{equation}
where $\oplus$ is the concatenation operation.

A simple trainable linear layer $g$, projects these video-level features into the language decoder’s embedding space, transforming them into corresponding language embedding tokens $Q_{v}$.
\begin{equation}
Q_v=g(R)\in\mathbb{R}^{(T+N)\times K}
\label{eq:3}
\end{equation}
where $K$ represents the dimension of the text query. 

Specific inquiries regarding emotional causes for explanations: User Query is denoted as $Q_t\in\mathbb{R}^{L\times K}$ where $L$  is the length of the query. These queries are tokenized to be dimensionally compatible with the video embeddings. Finally, $Q_v$ and $Q_t$ are input to the language decoder together.

\subsection{Facial Emotion Module}

As shown in the "Facial Emotion" section of Figure~\ref{fig:fig5}, we further design a two-stage system, where the first stage performs face detection to segment faces from the video, and the second stage conducts facial emotion recognition. This two-stage detection system reduces the impact of the environment on facial emotion recognition and improves the accuracy of detection.

In the first stage, in order to efficiently extract features and adapt to various scale changes in face, we use CNN \cite{lecun1998gradient} to extract feature maps $F$ from the input video frame $I$, and then, a set of multi-scale anchors is generated on the feature map, and these anchors $a$ are classified using Detection Layers composed of VGG16 \cite{simonyan2014very} and CNNs to determine whether they contain a face. Simultaneously, bounding box regression is applied to adjust the position and size of the candidate detection boxes. The process of classification and regression can be represented as:
\begin{equation}
\label{eq:eq4}
    \mathds{1}_{face},\ \Delta  = DetectionLayers(F,\ a)
\end{equation}
where $\mathds{1}_{face}$ represents whether it contains a face, and $\Delta$ represents the bounding box score. Using the regression output $\Delta$ to adjust the position and size of each anchor, $a$, the detection box, $B$, is generated. To reduce the probability of detecting the same face multiple times, we use Non-Maximum Suppression (NMS) \cite{felzenszwalb2009object} with an Intersection over Union (IoU) threshold to obtain the final detected faces.
\begin{equation}
\label{eq:eq5}
    B_{final}=\text{NMS}(B,\ \mathds{1}_{face},\ IoU_{threshold})
\end{equation}

\begin{table*}
\small
  \centering
    \caption{\label{tab:tab2}
    Comparison of Automatic Evaluation Metrics Experiments On ECEM.}
  \begin{tabular}{ccccccccc}
\toprule[1pt]
\textbf{Class} & \textbf{Method}       & \textbf{BLEU4}  & \textbf{METEOR} & \textbf{ROUGE}  & \textbf{BERTScore}&  \textbf{BLEURT}& \textbf{CIDEr} \\ \midrule[1pt]
\multirow{4}{*}{\textbf{Pre-trained Small Models}} & \textbf{IGV}         & 0.0082 & 0.1201	 & 0.2190 & 0.7920  & -1.5200 & 0.5146 \\
&\textbf{ObG}         & 0.0134 & 0.1791	 & 027880 & 0.8131  & -1.1456 & 0.5874 \\
&\textbf{COSA}         & 0.0064 & 0.1261	 & 0.2278 & 0.7674  & -1.3576 & 0.5710 \\
&\textbf{VALOR}         & 0.0085 & 0.1311	 & 0.2581 & 0.7964  & -1.2127 & 0.5264 \\ \hline
\multirow{8}{*}{\textbf{LLMs}}& \textbf{Qwen}         & 0.0105 & 0.1432 & 0.1222 & 0.8498 & -1.0425 & 0.5342 \\
&\textbf{Qwen1.5}         & 0.1124 & 0.2775 & 0.3157 & 0.7693 & -0.9878 & 0.4759 \\
&\textbf{Qwen2}         & \textbf{0.1235} & 0.3221 & 0.4045 & 0.8685 & -0.6177 & 0.6638 \\
&\textbf{Baichuan2}         & 0.0245 & 0.1354 & 0.1577 & 0.8246 & -1.1124 & 0.5344 \\
&\textbf{LLaMA2}       & 0.0106 & 0.1519 & 0.2443 & 0.8507 & -1.1445 & 0.5543 \\
&\textbf{LLaMA3}         & 0.0117 & 0.1499 & 0.2758 & 0.8617 & -0.9545 & 0.5698 \\
&\textbf{ChatGLM3}     & 0.0007 & 0.1857 & 0.1258 & 0.8587 & -0.7099 & 0.6049 \\
&\textbf{ChatGLM4}         & 0.0010 & 0.1694 & 0.1294 & 0.8629 & -0.6921 & 0.6196 \\ \hline
\multirow{8}{*}{\textbf{MLLMs}}& \textbf{ChatUniVi}    & 0.0136 & 0.1331 & 0.1501 & 0.8245 & -1.1092 & 0.4752 \\
&\textbf{Qwen2-VL}  & 0.0861 & 0.3214 & 0.3634 & 0.8721 & -0.5463 & 0.6557\\
&\textbf{NExT-GPT}     & 0.0945 & 0.1250 & 0.3401 & 0.7599 & -1.2821 & 0.6397 \\
&\textbf{Video-LLaMA}  & 0.0871 & 0.3324 & 0.3724 & 0.8981 & -0.5463 & 0.6677\\
&\textbf{Video-LLaMA2} & 0.0012 & 0.1785 & 0.1300 & 0.8361 & -0.8779 & 0.6268\\
&\textbf{LLaVA-NeXT-Video}  & 0.0654 & 0.3235 & 0.3966 & 0.8974 & -0.6634 & 0.6788 \\
&\textbf{FAME-Net (ours)}  & 0.0870 & \textbf{0.3423} & \textbf{0.4051} & \textbf{0.9053} & \textbf{-0.4789} & \textbf{0.6836} \\ 
\bottomrule[1pt]
  \end{tabular}
\end{table*}

\begin{table*}[t]
\small
  \centering
  \caption{\label{tab:tab999}
    Comparison of Automatic Evaluation Metric Experiments On ECGC.}
  \begin{tabular}{cccccccccc}
\midrule[1pt]
\textbf{Modality} &\textbf{Method}            & \textbf{BLEU-1} & \textbf{BLEU-2} & \textbf{BLEU-3} & \textbf{BLEU-4} & \textbf{METEOR} & \textbf{ROUGE-L} & \textbf{CIDEr}  & \textbf{BERTScore} \\ \midrule[1pt]
\multirow{4}{*}{\textbf{Text}}&\textbf{GPT-3.5}           & 0.2413          & 0.1582          & 0.1157          & 0.0886          & 0.1722          & 0.2258           & 0.8518          & 0.6801           \\
&\textbf{Gemini-Pro}        & 0.2624          & 0.1726          & 0.1272          & 0.0984          & 0.1767          & 0.2497           & 0.8926          & 0.6936           \\
&\textbf{T5s}                & 0.4815          & 0.4435          & 0.3718          & 0.3403          & 0.2979          & 0.4608           & 2.8770          & 0.7619           \\
&\textbf{Flan-T5}           & 0.4897          & 0.4212          & 0.3804          & 0.3499          & 0.3057          & 0.4751           & 3.0279          & 0.7698           \\ \hline
\multirow{4}{*}{\textbf{MM}}&\textbf{Gemini-Pro-Vision} & 0.2780          & 0.1826          & 0.1371          & 0.1085          & 0.1798          & 0.2453           & 0.8960          & 0.6960           \\
&\textbf{ObG}               & \textbf{0.5011} & \textbf{0.4341} & \textbf{0.3939} & \textbf{0.3641} & 0.3008          & 0.4712           & 3.0079          & 0.7672           \\
&\textbf{FlanObG}           & 0.4967          & 0.4313          & 0.3924          & 0.3631          & 0.3042          & 0.4781           & 3.0808          & 0.7711           \\
&\textbf{FAME-Net (ours)}   & 0.4578          & 0.4214          & 0.3921          & 0.3587          & \textbf{0.3150} & \textbf{0.4786}  & \textbf{3.1347} & \textbf{0.8454}  \\
\bottomrule[1pt]
\end{tabular}
\end{table*}

In the second phase, to reduce the loss of spatial information and retain more contextual details, we first use a MFN \cite{zadeh2018memory} to extract feature maps $F_{\mathrm{MFN}}$ from the input face $B_{final}$:
\begin{equation}
\label{eq:eq5}
    F_{\mathrm{MFN}}=\mathrm{MFN}(B_{final})
\end{equation}
The features of facial emotions are not limited to a single direction. By considering both directions simultaneously, the model can more comprehensively understand the complexity of facial expressions. Therefore, we use DDA \cite{zhang2023dual} to generate attention maps in both the vertical direction ($Y$) and the horizontal direction ($X$).
\begin{equation}
\begin{aligned}
    F_{\mathrm{DDA}}^{X} = \mathrm{DDA}_{X}(F_{\mathrm{MFN}}) \\
    F_{\mathrm{DDA}}^{Y} = \mathrm{DDA}_{Y}(F_{\mathrm{MFN}})
\end{aligned}
\end{equation}
We integrate attention mechanisms from two directions to enhance recognition capability and accuracy. The sigmoid function, $\sigma$, is used to suppress noise while strengthening the critical areas of the features. This result is multiplied by the max-pooling output to further enhance important features and suppress less significant information, yielding the final attention feature map $F_{att}$.
\begin{equation}
\begin{aligned}
    F_{\mathrm{att}} = pooling(F_{\mathrm{DDA}}^{X}, F_{\mathrm{DDA}}^{Y}) \cdot \sigma(F_{\mathrm{DDA}}^{X} F_{\mathrm{DDA}}^{Y})
\end{aligned}
\end{equation}
Finally, the attention feature map is passed through a fully connected layer to predict the emotion category of the face, $E_{cls}$.
\begin{equation}
\begin{aligned}
    E_{cls} = FC(F_{\mathrm{att}})
\end{aligned}
\end{equation}

\subsection{Video Instruction Tuning}
During the fine-tuning phase, we use predefined prompts, \texttt{Prompt}$_{\text{Video}}$.

\begin{description}
    \item  \colorbox{gray!20}{\parbox{1.0\linewidth}{\small\texttt{USER: <Instruction> <User Query><Vid-tokens><Emotion>\\
    Assistant: Using the notations, we can represent it as:\\
    USER: <$Q_t$> <$Q_v$> <$E_{cls}$>\\
    Assistant: }}}
\end{description}
where <Instruction> defines the task, which requires explaining the emotional reasons by understanding the video. <User Query> includes the dialogue history in multimodal conversations, as well as the utterance requiring emotional cause explanation and the emotion it conveys. The predicted answer corresponds to the final cause $C$. Throughout the training process, the weights of the video encoder and LLM remain freezed, and the model maximizes the likelihood of tokens representing the answer by adjusting the linear layer. <$Q_t$> represent <Instruction> and <User Query>; <$Q_v$> represent <Vid-tokens>.

\section{Experiments}

\subsection{Implementation Details}

Our FAME-Net model is built on the LLaVA-7b \cite{NEURIPS2023_6dcf277e} foundation. To enhance training efficiency, we preprocess the faces in the video before training. The training process is conducted on two NVIDIA A6000 48G GPUs.

\subsection{Evaluation Metrics}
To comprehensively evaluate the fluency, diversity, and accuracy of generated causeing, we adopt a series of widely recognized automatic evaluation metrics for text generation, including BLEU \cite{papineni2002bleu}, METEOR \cite{banerjee2005meteor},  ROUGE \cite{lin2004rouge}, and CIDEr \cite{vedantam2015cider}, which focus on lexical and syntactic matching, as well as BERTScore \cite{zhang2019bertscore} and BLEURT \cite{sellam2020bleurt}, which assess whether the generated text is semantically consistent with the reference text.

\subsection{Baselines}

We compare it with several open-source models of similar scale. Ultimately, we select NExT-GPT \cite{wu2023next}, ChatUniVi \cite{jin2023chatunivi}, Video-LLaMA\cite{zhang2023video}, Qwen2-VL \cite{wang2024qwen2}, LLaVA-NeXT-Video \cite{liu2024llavanext}, and Video-LLaMA2 \cite{cheng2024videollama} as baselines. Also, we choose models like Qwen \cite{bai2023qwen}, Qwen1.5 \cite{bai2023qwen}, Qwen2 \cite{qwen2}, Baichuan2 \cite{yang2023baichuan}, LLaMA2 \cite{touvron2023llama}, LLaMA3 \cite{grattafiori2024llama}, and ChatGLM3 \cite{chatglm3}, ChatGLM4 \cite{glm2024chatglm}, which excel in single text modality, for experiments where only conversation history is input. In addition, we select four pre-trained small models as baselines. The four small models are VALOR \cite{liu2024valor}, COSA \cite{chen2023cosa}, ObG \cite{wang2024observe}, and IGV \cite{li2022invariant}. We train and test these models on the ECEM dataset. For the ECGC dataset, we use the experimental results from the original paper as baselines.

\subsection{Main Results and Analysis}

\subsubsection{Automatic Evaluation}
\begin{table*}[h]
\small
  \centering
    \caption{\label{tab:tab3}
    Ablation Study Results, where V represents the visual modality, T represents the text modality, and E represents the facial emotion recognition module.}
\begin{tabular}{cccccccc}
\toprule[1pt]
\textbf{Method}       & \textbf{BLEU4}  & \textbf{METEOR} & \textbf{ROUGE}  & \textbf{BERTScore}& \textbf{BLEURT}& \textbf{CIDEr}           \\ \midrule[1pt]
\textbf{V}      & 0.0375         & 0.2217          & 0.2455          & 0.8850          & -0.9061          & 0.5572          \\
\textbf{V+E}    & 0.0695         & 0.3058          & 0.3427          & 0.8960          & -0.7715          & 0.6259          \\
\textbf{V+T}    & 0.0700        & 0.3233          & 0.3771          & 0.8977          & -0.7215          & 0.6626          \\
\textbf{V+T+E} & \textbf{0.0870} & \textbf{0.3423}          & \textbf{0.4051} & \textbf{0.9053}   & \textbf{-0.4789} & \textbf{0.6836} \\ 
\bottomrule[1pt]
\end{tabular}
\end{table*}

\begin{table*}[h]
\small
\centering
\caption{\label{tab:tab7}
    Comparison between original data and transcription results.}
\begin{tabular}{cccccccc}
\toprule[1pt]
\textbf{Method}       & \textbf{BLEU4}  & \textbf{METEOR} & \textbf{ROUGE}  & \textbf{BERTScore}& \textbf{BLEURT}& \textbf{CIDEr}  \\ \midrule[1pt]
\textbf{Original Text}          & 0.0495         & \textbf{0.3425} & 0.4040          & 0.8978            & -0.5031          & 0.6755          \\
\textbf{Improved Text} & \textbf{0.0870} & 0.3423          & \textbf{0.4051} & \textbf{0.9053}   & \textbf{-0.4789} & \textbf{0.6836} \\ \bottomrule[1pt]
\end{tabular}
\end{table*}

Table~\ref{tab:tab2} presents the performance of the baseline model and FAME-Net on the ECEM dataset across six evaluation metrics. From the table, it is evident that the FAME-Net model excels in multiple key metrics, particularly achieving the best performance in METEOR, ROUGE, BERTScore, BLEURT, and CIDEr. These metrics outperform other comparative models, including pre-trained small models, Large Language Models (LLMs), and Multimodal Large Language Models (MLLMs). Although FAME-Net does not achieve the highest score in the BLEU4 metric, this does not detract from its overall superior performance. BLEU4, as an evaluation metric based on n-gram overlap, can reflect the local accuracy of generated text in some cases, but its ability to capture semantic understanding and contextual coherence is limited. In contrast, other metrics place more emphasis on semantic similarity and contextual consistency. Therefore, the excellent performance of FAME-Net in these metrics fully demonstrates its strong capability in the task of MECEC, particularly highlighting its significant advantages in semantic understanding and generation quality.

Table~\ref{tab:tab999} shows the scores of the FAME-Net model on the ECGC dataset, where GPT-3.5 \cite{OpenAI2023}, Gemini-Pro \cite{team2023gemini}, T5 \cite{raffel2020exploring}, and Flan-T5 \cite{raffel2020exploring} are pure text models, and Gemini-Pro-Vision \cite{team2023gemini}, ObG \cite{wang2024observe}, and FlanObG \cite{wang2024observe} are multimodal models. The experimental results show that FAME-Net performs weakly on the BLEU metric but leads on metrics such as METEOR, ROUGE-L, CIDEr, and BERTScore. This indicates that our model has superior capabilities in semantic understanding, fluency, and creativity. Since the BLEU metric mainly focuses on n-gram matching, it may not fully reflect the model's advantages in language generation, while other metrics emphasize the semantic consistency and contextual understanding of the text. Therefore, despite the lower BLEU score, the model's outstanding performance in generation quality and expressive ability still indicates its higher potential in practical applications.

Overall, the FAME-Net model performs excellently on ECGF and ECEM datasets, showing leading advantages in multiple metrics and achieving a good balance between innovation and accuracy. In addition, both datasets demonstrate that multimodal models, including FAME-Net, outperform pure text models and other smaller models in overall performance. This indicates that integrating information from multiple models enhances the model's ability to solve MECEC task.

\subsubsection{Human Evaluation and Analysis}
In addition to automatic evaluation, we randomly select 100 test samples from the four best-performing models for manual evaluation. We choose three graduate students specializing in sentiment analysis as evaluators and ask them to score the generated reasons based on three criteria: coherence, fluency, and relevance, with a rating scale from 1 to 5. Coherence emphasizes the internal consistency of generated text in terms of logic, semantics, and structure. Fluency evaluates the grammatical correctness and naturalness of language use in the sentences. Relevance determines whether the generated sentences contain the exact reasons for expressing emotions. As shown in Table~\ref{tab:tab4}, FAME-Net achieves the best performance across all three metrics. This result demonstrates that FAME-Net not only excels in the logical consistency and linguistic naturalness of generated text but also more accurately captures the reasons related to emotional expression, showcasing its strong capability in the MECEC task.

\subsection{Ablation Experiments}
The ablation experiment results, as shown in Table~\ref{tab:tab7}, confirm that multimodal information fusion enhances dialogue generation performance. The joint visual-text modeling improves the average performance across all metrics by 7.1\% compared to the pure visual model, demonstrating the effectiveness of cross-modal semantic alignment. The introduction of the facial emotion module increases METEOR and BLEURT by 2.5\% and 7.3\%, respectively, indicating that emotional features can enhance the empathetic ability of expressions. The final V+T+E model achieves optimal performance across all evaluation metrics, validating the synergistic effect of visual content understanding, textual semantic reasoning, and emotional feature capturing. This proves that complementary multimodal information can more comprehensively support dialogue generation tasks.

\subsection{Original Data and Transcription}

To verify our hypothesis that audio transcription data is of higher quality compared to the original text data, we conduct a detailed and thorough comparative experiment, with the results comprehensively displayed in Table~\ref{tab:tab7}. The BLEU4 score is higher for the transcription, indicating that the transcribed text aligns more closely and accurately with the reference standard in terms of lexical matching, thereby demonstrating superior accuracy and consistency. Although the METEOR score remains nearly identical between the two datasets, the transcription results still slightly outperform the original text across all other evaluation metrics, suggesting an overall enhancement and refinement in data quality following transcription.

\begin{table}[h]
\small
\centering
\caption{\label{tab:tab4}
    Human evaluation results, Coh, Flu, Rel represents Coherence, Fluency, Relevance.}
\begin{tabular}{cccc}
\toprule[1pt]
\textbf{Method}       & \textbf{Coh} & \textbf{Flu}       & \textbf{Rel}     \\ \midrule[1pt]
\textbf{LLaVA-NeXT-Video} & 3.91 & 4.14          & 3.85          \\
\textbf{Video-LLaMA}  & 4.05          & 4.24          & 3.99          \\
\textbf{Video-LLaMA2} & 3.97          & 4.17          & 3.89          \\
\textbf{FAME-Net (ours)}  & \textbf{4.15}          & \textbf{4.31} & \textbf{4.01} \\ \bottomrule[1pt]
\end{tabular}

\end{table}

\section{Conclusion}
Our paper constructs a novel dataset, ECEM, for training and evaluating the MECEC task. To address the MECEC task, we propose a specialized model, FAME-Net, based on the LLaVA framework. FAME-Net excels in understanding video content and provides natural language explanations for the emotions of characters depicted in the videos. Extensive experimental results demonstrate that our model performs exceptionally well in the MECEC task. Specifically, FAME-Net, by integrating multimodal information, captures emotional changes in characters more accurately and generates coherent and emotionally rich explanations. Additionally, the construction of the ECEM dataset considers diversity and complexity, covering various emotion types and scenarios, ensuring the model's generalization ability in practical applications.

\section*{ACKNOWLEDGEMENTS}
This work is supported by the National Natural Science Foundation of China (62272092,62172086) and the Fundamental Research Funds for the Central Universities of China (No.N2116008).

\bibliographystyle{ACM-Reference-Format}
\balance
\bibliography{sample-base}

\end{document}